\documentclass{article}

\PassOptionsToPackage{numbers, compress}{natbib}
\usepackage[preprint]{neurips_2025}

\usepackage[utf8]{inputenc} % allow utf-8 input
\usepackage[T1]{fontenc}    % use 8-bit T1 fonts
\usepackage[colorlinks=true, linkcolor=blue, citecolor=blue, urlcolor=blue]{hyperref}
\usepackage{url}            % simple URL typesetting
\usepackage{booktabs}       % professional-quality tables
\usepackage{amsfonts}       % blackboard math symbols
\usepackage{nicefrac}       % compact symbols for 1/2, etc.
\usepackage{microtype}      % microtypography
\usepackage{xcolor}         % colors
\usepackage{amssymb}
\usepackage{amsmath}
\usepackage{graphicx} 
\usepackage{float} 

\usepackage{etoolbox}

\usepackage{algorithm}
\usepackage{algpseudocode}
\algrenewcommand\algorithmicrequire{\textbf{Input:}}
\algrenewcommand\algorithmicensure{\textbf{Output:}}

\usepackage{placeins}  % adds \FloatBarrier
\usepackage{subcaption}                 % for subfigures
\usepackage[font=small]{caption}        % slightly smaller captions
\title{Continual Learning with Query-Only Attention}

\author{%
  Gautham Bekal \\
  Mitchell, Enlyte \\
  \texttt{gauthambekal93@gmail.com}
  \And
  Ashish Pujari \\
  Department of Mechanical Engineering \\
  University of North Carolina at Charlotte \\
  \texttt{apujari1@uncc.edu}
  \And
  Scott David Kelly \\
  Department of Mechanical Engineering \\
  University of North Carolina at Charlotte \\
  \texttt{skelly52@charlotte.edu}
}

\begin{document}

\maketitle

\begin{abstract}
Continual learning involves learning from a stream of data without repetition of data points, a scenario that is inherently complex due to distributional shift across tasks.
We propose a query-only attention mechanism that discards keys and values, yet preserves the core inductive bias of full-attention architectures. In continual learning scenarios, this simplified mechanism significantly mitigates both loss of plasticity and catastrophic forgetting, outperforming baselines such as selective re-initialization. We establish a conceptual link between query-only attention, full transformer attention, and model agnostic meta-learning, framing them as instances of meta-learning. We further provide intuition for why query-only attention and full-attention networks help preserve plasticity in continual settings. Finally, through preliminary Hessian spectrum analysis, we observe that models maintaining higher curvature rank across tasks tend to retain plasticity. Our findings suggest that full attention may not be essential for capturing the benefits of meta-learning in continual learning.
\end{abstract}

\section{Introduction}
Continual learning remains a fundamental challenge in deep learning,\citep{wang2024comprehensive} where models must learn from non-stationary data streams without succumbing to catastrophic forgetting \citep{kirkpatrick2017overcoming} or loss of plasticity \citep{dohare2024loss} . While many existing approaches mitigate forgetting using replay buffers \citep{zhang2024core}, regularization, or selective re-initialization, preserving plasticity—the capacity to adapt to new tasks—remains significantly more elusive.

Recent work has shown that attention network \citep{vaswani2017attention} in transformer architectures, originally developed for sequence modeling, exhibit strong performance in continual learning . Notably, attention networks tend to retain plasticity across tasks more effectively than traditional MLPs or CNNs \citep{wang2025replay}. Motivated by this observation, we ask: Can the core mechanism of transformers—attention—be further simplified while retaining its continual learning benefits?

To this end, we propose a query-only attention mechanism that removes keys and values from the attention layer. Surprisingly, this minimalist design not only retains the benefits of full attention in continual learning but often surpasses it. Our experiments show that query-only attention significantly reduces both catastrophic forgetting and plasticity loss, without relying on task boundaries or selective re initialization.

Moreover, we show a deeper relationship between query-only attention, full attention network and  Model-Agnostic Meta-Learning (MAML) \citep{finn2017maml}. Through empirical comparisons and Hessian-based curvature analysis, we demonstrate that both our model and MAML maintain stable near constant hessian ranks throughout training—a hallmark of models that preserve plasticity and enable rapid adaptation.

While much of continual learning research has focused on catastrophic forgetting, we emphasize that our primary goal is mitigating loss of plasticity—the declining ability of models to acquire new knowledge. In our framework, reduced forgetting emerges as a natural byproduct of improved plasticity, rather than the central objective.

Our key findings are:
\begin{itemize}
\item We introduce \textit{Query-Only Attention}, which is based on attention mechanism and meta-learning that mitigates loss of plasticity more effectively than state-of-the-art methods in fully online continual learning experiments.
\item As a natural consequence, Query-Only Attention also mitigates catastrophic forgetting when task identity is available, although \textit{forgetting reduction is not the primary focus of this work}.
\item We provide a conceptual explanation showing that Query-Only Attention mitigates both loss of plasticity and forgetting by converging toward a \textit{global} rather than task-specific \textit{local} solution.
\item We establish conceptual links between Query-Only Attention, the original attention mechanism \citep{vaswani2017attention}, and meta-learning approaches such as MAML \citep{finn2017maml} in continual learning context.
\item We analyze the Hessian spectrum and effective rank \citep{lewandowski2023curvature}, demonstrating that decreasing rank across tasks correlates with loss of plasticity.
\end{itemize}

\section{Related work}
Deep neural networks have shown remarkable generalization capabilities on unseen tasks. However, they typically operate under the assumption that the training data is stationary and that all samples are available simultaneously during training. In contrast, online learning assumes that data arrives sequentially in a stream and each data point is observed only once, eliminating the concept of epochs.
In such a setting, the model must continuously update its parameters to adapt to incoming data. From the model’s perspective, the data distribution is inherently non-stationary, since not all samples are available at the same time. This leads to two major challenges: catastrophic forgetting and loss of plasticity.
Catastrophic forgetting — the degradation of performance on previously learned tasks after training on new ones — has been extensively studied in the literature \citep{mccloskey1989catastrophic}, \citep{kemker2018measuring}, \citep{ramasesh2021anatomy}.
A more fundamental and less studied issue is loss of plasticity, the gradual reduction in a model’s ability to learn new tasks altogether \citep{dohare2024loss}, \citep{mirzadeh2020understanding}

Hence, continual learning faces challenge on two fronts, forward performance / mitigating loss of plasticity and backward performance / mitigating catastrophic forgetting. \citep{chen2023stability} showed that
there exists a tradeoff between the two. We show an alternative view that 
Query-Only Attention and related architectures attain a global solution which mitigates both loss of plasticity and catastrophic forgetting simultaneously.

Most papers, analyze one of the two of above challenges, however very few papers tackle both challenges simultaneously.
Regularization-based continual learning methods such as Elastic Weight Consolidation (EWC) \citep{kirkpatrick2017overcoming}, \citep{zenke2017si}, and Learning without Forgetting \citep{li2016lwf} were designed primarily to mitigate catastrophic forgetting, i.e., the degradation of previously learned tasks. However, these approaches do not directly address the complementary challenge of loss of plasticity, where the model fails to acquire new tasks altogether. Our focus in this work is specifically on loss of plasticity, where forgetting is a downstream consequence rather than the central phenomenon. For this reason, we compare against baselines that are explicitly targeted at plasticity, including \citep{dohare2024loss} and recent attention-based approaches \citep{wang2025replay}, rather than against EWC-style methods that operate in an orthogonal regime.

One such paper is \citep{elsayed2024addressing} which uses utility based methodology for handling both catastrophic forgetting and loss of plasticity. Here, the authors are working with unknown task boundaries. However, obtaining the utility is expensive especially in the era of large and very 
large models. Controlling gradient updates based on weight utility leads to reduced ability to retain old tasks as their number increases. Most importantly, this method is ad-hoc solution for continual learning problem and not a more global solution which can scale to very large number of tasks.
Our method contrasts in that it obtains a global solution and thus has no reduction in performance irrespective number of tasks.

The core algorithm we developed is most closely related to the paper \citep{wang2025replay} which utilizes attention network and replay buffer to handle this challenge. However, attention network has O(n**2) in compute which can be challenging in continual learning setting where data will come rapidly.  Here, n is the size of replay buffer. 
We draw our inspiration from this paper on using replay buffer but make a novel hypothesis that query matrix is all thats needed for continual learning. Our method achieves similar or superior performance compare to full attention and can also do the compute in O(n).
The other aspect being \citep{wang2025replay}  does not explain the intriguing phenomenon.
Here we carry out a detailed empirical and theoretical analysis on why query only attention works.

Our work reveals  deep connection between attention network, in-context learning \citep{dong2024survey}, \citep{wu2025why}, \citep{ahn2023transformers} model agnostic meta learning \citep{finn2017maml}, and metric based meta learning \citep{vinyals2016matching}, \citep{koch2015siamese}, \citep{snell2017prototypical} under the paradigm of continual learning.

To understand the mechanism of loss of plasticity in continual learning, we utilize a robust metric of calculating effective rank of hessian matrix as shown in \citep{lewandowski2023curvature}, \citep{roy2007effective}.
They show that maintaining plasticity requires the effective rank to remain high rather than decreasing.
In our study we pair it with our algorithm of query only attention to justify its efficiency in mitigating loss of plasticity by maintaining a non decreasing effective rank across continual learning setting.

\section{Background and preliminaries}

Our theoretical analysis builds on several standard components: attention networks, 
meta-learning (in particular MAML), and $k$-nearest neighbors (KNN). 
We briefly review each here to fix notation and highlight the connections 
that will be used in Section~\ref{sec:theory}.

\subsection{Attention mechanism}

In the standard attention network \citep{vaswani2017attention}, each query vector $q_i$ produces a weighted combination of value vectors $V = \{v_1, \dots, v_n\}$:  

\begin{equation}
\text{Attn}(q_i, K, V) \;=\; \sum_{j=1}^n \alpha_{ij} \, v_j,
\label{eq:attention}
\end{equation}

where the weights $\alpha_{ij}$ are obtained from a softmax over query–key dot products:  

\begin{equation}
\alpha_{ij} \;=\; \frac{\exp\big( q_i^\top k_j / \sqrt{d} \big)}
{\sum\limits_{j'=1}^n \exp\big( q_i^\top k_{j'} / \sqrt{d} \big)} .
\label{eq:attention_weights}
\end{equation}

Here $K = \{k_1, \dots, k_n\}$ are key vectors and $d$ is the feature dimension.  
This requires computing all pairwise dot products $q_i^\top k_j$, which scales as $\mathcal{O}(n^2)$ in sequence length $n$.  
In contrast, our query-only model removes $K$ and $V$, learning task-specific weights $\theta'$ directly, while still preserving the interpretation of predictions as weighted combinations over context points.

\subsection{Meta-learning}
Meta-learning aims to enable rapid adaptation to new tasks from a few support examples \citep{finn2017maml}. 
Traditional meta-learning assumes task boundaries and episodic training, and is thus non-continual. 
Recent work \citep{javed2019meta} , \citep{son2023meta} explores meta-learning for continual learning. 
Our approach draws inspiration from meta-learning while operating fully online.

\paragraph{Model-agnostic meta-learning}\hfill \\ 
MAML \citep{finn2017maml} optimizes model parameters through inner- and outer-loop updates:  

\begin{equation}
\Theta'_i = \Theta - \alpha \nabla_\Theta \mathcal{L}_\Theta(\text{support}_i)
\label{eq:maml_inner}
\end{equation}

\begin{equation}
\Theta \leftarrow \Theta - \beta \sum_{i=1}^m \nabla_\Theta \mathcal{L}_{\Theta'_i}(\text{query}_i).
\label{eq:maml_outer}
\end{equation}

\subsection{$k$-Nearest neighbors (KNN)}
In $k$-nearest neighbors regression, prediction is based on the $k$ closest 
points in a support set $\mathcal{S} = \{(x_i,y_i)\}_{i=1}^N$ under a distance 
metric $d(\cdot,\cdot)$ (e.g., Euclidean).  

Given a query input $x$, let $\mathcal{N}_k(x)$ denote the indices of the $k$ 
nearest neighbors. The kNN prediction is the local average:
\begin{equation}
\hat{y}(x) = \frac{1}{k} \sum_{i \in \mathcal{N}_k(x)} y_i.
\label{eq:knn}
\end{equation}

Thus, kNN regression is a non-parametric, memory-based method where predictions 
adapt directly from nearby support examples.

\subsection{In-context learning}
In-context learning has recently been interpreted as an implicit $k$-nearest-neighbors mechanism that emerges in the forward pass of transformers \citep{olsson2022induction}. 
This perspective provides a key motivation for our work: by modifying attention, we aim to enable continual adaptation without requiring explicit task identifiers.

\section{Problem statement}
We study the \textit{continual learning (CL)} setting, where a model receives a continuous stream of data. Each data point is observed \textit{once} during training, without repetition or epochs.  

Formally, the stream is generated from a sequence of tasks 
\(\{T_1, T_2, \dots, T_n\}\). Each task \(T_i\) is associated with a (potentially non-stationary) distribution \(\mathcal{D}_i(x,y)\) over input–label pairs \((x,y)\).  

\textbf{Training.}
\begin{itemize}
    \item The model does not observe task boundaries or task identities.
    \item Samples arrive sequentially, drawn from the evolving distribution.
    \item The objective is to update the model online while retaining performance across all tasks.
    \item Based on the task at hand, the model may update on a single data point or a batch of data-points.
\end{itemize}

\noindent
\textbf{Evaluation Protocol.} 
During inference, the model processes a data stream sequentially. 
At data point \(m\) of task \(t\), the goal is to predict the next \(n\) points 
\(\{m+1, \dots, m+n\}\) from the same task. 
The resulting accuracy (or loss) defines the \textit{forward performance}; 
its degradation over time indicates \textit{loss of plasticity}. 

After training up to task \(t\), the model is also evaluated on samples from a previous task \(t-j\). 
The resulting accuracy defines the \textit{backward performance}, 
and its degradation quantifies \textit{catastrophic forgetting}. 

\begin{itemize}
    \item \textbf{Forward testing (plasticity):} 
    Evaluates on upcoming data from the current stream without task identifiers. 
    A decline in this metric across tasks signals loss of plasticity.
    
    \item \textbf{Backward testing (forgetting):} 
    Evaluates on a small held-out buffer of past-task samples where task identities are known. 
    A decline in this metric indicates catastrophic forgetting.
\end{itemize}

\section{Method}
\subsection{Query only model with replay buffer}
\label{sec:method}

Drawing the connection from  attention networks, meta-learning, KNN and replay buffer we design an architecture which obtains an optimal solution for continual learning task, leading to mitigation 
of both loss of plasticity and catastrophic forgetting simultaneously.
A sample data point is $d = (x, y) \in \mathcal{D}$.
Let, $x \in \mathbb{R}^a$ and  $y \in \mathbb{R}^b$
We define a buffer $\mathcal{B}$ containing $n$ data-points.
Every time step we construct a support set $\mathcal{S}$ of size $m$ sampled from $\mathcal{B}$ such that $m<=n$ depending on the problem at hand.
\\Hence,  $S \in \mathbb{R}^{m* (a+b)}$ 

\[
S =
\begin{bmatrix}
x_{s1} & y_{s1} \\
x_{s2} & y_{s2} \\
\vdots & \vdots \\
x_{sm} & y_{sm}
\end{bmatrix}
=
\begin{bmatrix}
s_1 \\
s_2 \\
\vdots \\
s_m
\end{bmatrix}
\]
\\Let, the data-point on which we want to make prediction be query $q_i$, such that $q_i=x_i$.
The neural-net model is a single query-matrix $\mathcal{Q_\theta} \in \mathbb{R}^{(2a + b)*b}$, for illustration purpose and can have multiple layers.

\begin{algorithm}[ht]
\caption{Query-Only Attention with Replay Buffer}
\label{alg:query_only}
\begin{algorithmic}[1]
\Require Stream of tasks $\{T_1, T_2, \ldots\}$; replay buffer $\mathcal{B}$ of size $n$; support size $m$; query-only model $Q_\theta$; learning rate $\eta$
\Ensure Updated parameters $\theta$
\State Initialize $\theta$, $\mathcal{B} \gets \emptyset$
\For{each incoming sample $(x_t,y_t)$}
    \State Insert $(x_t,y_t)$ into buffer $\mathcal{B}$; evict oldest if $|\mathcal{B}| > n$
    \State Sample support set $\mathcal{S} = \{(x_j,y_j)\}_{j=1}^m \subset \mathcal{B}$
    \State Compute scores $d_j \gets Q_\theta(x_t, x_j, y_j)$
    \State Prediction $\hat{y}_t \gets \sum_{j=1}^m d_j \, y_j$
    \State Loss $\mathcal{L}_t \gets \ell(\hat{y}_t, y_t)$
    \State Update $\theta \gets \theta - \eta \nabla_\theta \mathcal{L}_t$
\EndFor
\end{algorithmic}
\end{algorithm}

We can thus write the predictive model as,
\begin{equation}
\hat{y_t}(x) = {\sum\limits_{x_i , y_i\in B} Q_\theta(x_t, x_i, y_i) * y_i}
\label{eq:query_based_model}     
\end{equation}
Here, $x_t$ is the query point, $x_i$ is a support input, and $y_i$ is support label.
$Q_\theta(x_t, x_i, y_i)$ denotes a learned similarity (or distance) function.
Unlike standard metrics such as dot products or Euclidean distance—which require vectors to be in the same feature space—$Q_\theta$ learns a representation where query–support pairs become directly comparable, allowing flexible weighting even when raw dimensions differ.

\subsection{MAML with replay buffer}
\paragraph{Note.} 
The adaptation of MAML with replay buffer is still work in progress. 
We include it here to illustrate a potential direction for combining meta-learning with large 
buffers in continual learning, but do not claim it as a finalized or fully validated algorithm. 
That said, our preliminary results are promising for one of the experiments, and suggest this variant may provide 
a complementary approach to attention-based or query-only models.  

We adapt MAML to the continual learning setting by introducing a large replay buffer. 
To minimize interference across tasks, the buffer is evenly partitioned into $t$ sub-buffers, 
one per sampled task. Without such partitioning, the sampled support and query examples 
from different tasks would overlap excessively, leading to degraded task separation 
and unstable meta-updates. Each task provides a small support set ($s$) and query set ($q$), 
which is sufficient for the MAML inner/outer loops.  

An important property of this setup is that the support/query sizes remain much smaller 
than those required by attention-based or query-only models. This makes MAML particularly 
well-suited for backward evaluation (catastrophic forgetting), since only a small set of 
examples per task is needed to adapt. Standard MAML inner-loop adaptation and outer-loop 
meta-updates are then applied on these partitioned tasks.  
The details of the algorithm are presented in the appendix section.

\begin{algorithm}[t]
\caption{MAML with Replay Buffer (work in progress)}
\label{alg:maml_replay}
\begin{algorithmic}[1]
\Require Stream $(x_t,y_t)$; replay buffer $\mathcal{B}$ of size $N$; 
tasks $t$; support $s$; query $q$; learning rates $\alpha,\beta$
\State Initialize $\theta$, $\mathcal{B} \gets \emptyset$
\For{each incoming $(x_t,y_t)$}
    \State Insert $(x_t,y_t)$ into $\mathcal{B}$; evict oldest if $|\mathcal{B}|>N$
    \State Partition $\mathcal{B}$ into $t$ equal sub-buffers
    \For{each task $i=1\dots t$}
        \State Sample $s$ support and $q$ queries from sub-buffer $i$
        \State Inner update on support with step size $\alpha$
        \State Outer update on queries with step size $\beta$
    \EndFor
\EndFor
\end{algorithmic}
\end{algorithm}

\section{Theoretical discussion}
\label{sec:theory}

\subsection{Global vs local solution}
\label{sec:local_global}
To understand why our algorithm overcomes loss of plasticity and catastrophic forgetting, we
first define weighted $k$-nearest neighbors

\begin{equation}
\hat{y}(x) = 
\frac{\sum\limits_{i \in N_k(x)} w(x, x_i)\, y_i}
     {\sum\limits_{i \in N_k(x)} w(x, x_i)},
\label{eq:weighted_knn}     
\end{equation}

where $w(x, x_i)$ is a weight function that decreases as the distance $d(x, x_i)$ increases.  
Common choices include:

\begin{align}
w(x, x_i) &= \frac{1}{d(x, x_i)}, \\
w(x, x_i) &= e^{-\alpha\, d(x, x_i)}.
\end{align}

In Equation~\ref{eq:weighted_knn}, predictions depend only on neighboring data points and not on any learnable parameters.  
Thus, weighted $k$NN avoids loss of plasticity (no parameters to get stuck in low-rank regions) and catastrophic forgetting (no parameters to overwrite).  
Performance is fully determined by the support set and chosen distance metric.  

Comparing Equation~\ref{eq:weighted_knn} with Equation~\ref{eq:query_based_model}, the difference lies in how the distance metric is obtained: in $k$NN it is fixed manually, while in the query-only attention it is learned as $\theta$.  
Once $\theta$ is learned, predictions depend primarily on the support set, so continual adaptation occurs in-context rather than through constant parameter updates.  
This makes the learning global in nature and independent of any single task which is different from vanilla backpropagation or continual backpropagation algorithm where the model updates its parameters continuously local for each task.

%\clearpage

\subsection{Model agnostic meta-learning for continual learning}
We can rewrite equation~\ref{eq:query_based_model} as,

\begin{equation}
\hat{y}_t(x) = \sum_{x_i, y_i \in B} \theta'_{t,i} \, y_i,
\label{eq:query_based_model_2}
\end{equation}

From Equation~\ref{eq:query_based_model_2}, predictions depend on task-specific parameters $\theta'$ generated at inference time.
This parallels the task-specific adaptation in MAML’s inner/outer-loop updates (Equations~\ref{eq:maml_inner}, \ref{eq:maml_outer}).
However, unlike query-only attention models, MAML was originally designed with task IDs and distributions known, an assumption that breaks in continual learning.
In our experiments, we find that using a large replay buffer stabilizes MAML training despite this limitation.
Even more interestingly, MAML requires a much smaller support set than query-only attention or full attention networks, which could make it a promising direction for mitigating catastrophic forgetting in future continual learning work where memory efficiency is crucial.

\subsection{Relationship to attention network}
From Equation~\ref{eq:attention}, the prediction is a linear combination of value vectors, which are themselves transformed representations of the input.  
Comparing this with Equation~\ref{eq:query_based_model_2}, we see a strong similarity: both aggregate information from a support set using learned weights.  
The main difference is that in attention (Equation~\ref{eq:attention_weights}), task specific weights are derived from query–key dot products, whereas in the query-only model they come from a learned distance metric $\theta'$.  

This equivalence suggests that attention networks can also mitigate loss of plasticity, much like the query-only model.  
However, computing attention weights requires all pairwise query–key dot products, leading to $\mathcal{O}(n^2)$ complexity for a support set of size $n$.  
In contrast, our query-only model only compares the current query with the support set, reducing the complexity to $\mathcal{O}(n)$.  
This efficiency allows us to scale to larger support sets in continual learning, improving performance without incurring the prohibitive cost of full attention.

\subsection{Hessian rank analysis}

To study plasticity, we analyze the \textit{effective rank} of the Hessian \citep{lewandowski2023curvature},\citep{roy2007effective},  
defined as
\begin{equation}
\mathrm{erank}(H) = \exp\!\Big(-\sum_{i=1}^n p_i \log p_i\Big), 
\quad 
p_i = \frac{\lambda_i}{\sum_j \lambda_j},
\label{eq:effective_rank}
\end{equation}
where $\{\lambda_i\}$ are the eigenvalues of the Hessian.  
A stable effective non-decreasing effective rank indicates models that preserve plasticity across tasks.

% ---------- EXPERIMENTS SECTION ----------

%\FloatBarrier
\section{Experiments}
We evaluate forward (plasticity) and backward (forgetting) performance on three benchmarks: 
\textbf{Permuted MNIST} (abrupt shifts), \textbf{Tiny ImageNet}, and 
\textbf{Slowly Changing Regression (SCR)} (gradual drift).
The above experiments follow same setup as explained in \citet{dohare2024loss}, except instead of full ImageNet we choose Tiny ImageNet for efficiency.
Distinction is that we perform experiments in a online setting, using \emph{unknown} task boundaries to study loss of plasticity and \emph{known} task boundaries to study catastrophic forgetting.
Detailed configurations appear in the Appendix. 
Results are averaged over three seeds with shaded $\pm$1 std regions. 
Higher accuracy (classification) and lower MSE (regression) indicate better performance. 
Baselines include \textbf{BP}, \textbf{CBP}, and the \textbf{Full-Attention Network}.
Since the primary focus of this work is mitigating loss of plasticity, we additionally include the state-of-the-art forgetting baseline \textbf{Elastic Weight Consolidation (EWC)} in the ImageNet experiments for completeness.

% ---------- Permuted MNIST: forward + backward in one figure ----------
\subsection{Permuted MNIST}
We evaluate the \textbf{query-only attention} model with support sizes of $1000$ and $200$ and
name them as Query-Only Attention V1 and Query-Only Attention V2. 
The \textbf{full-attention} uses support $100$, since its $\mathcal{O}(n^2)$ attention limits scalability, while our $\mathcal{O}(n)$ query-only design allows larger supports at lower cost. 
The \textbf{MAML}-style model uses a large replay buffer with a small support of $10$, trained on $100$ tasks. 
Each iteration is costlier due to inner-loop updates, so it runs on fewer tasks but performs multiple updates per iteration.
For readability, performance curves are averaged over fixed task windows;  results without averaging are included in the Appendix.

\label{subsec:permuted-mnist}
\begin{figure}[H]
  \centering
  \begin{subfigure}[t]{0.48\linewidth}
    \centering
    \includegraphics[width=\linewidth]{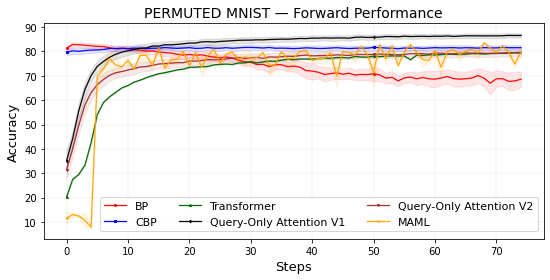}
    \caption{Forward accuracy (permuted MNIST).}
    \label{fig:mnist_fwd}
  \end{subfigure}\hfill
  \begin{subfigure}[t]{0.48\linewidth}
    \centering
    \includegraphics[width=\linewidth]{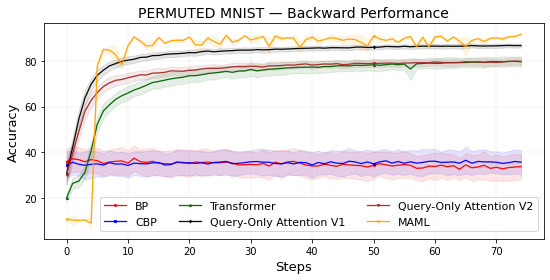}
    \caption{Backward accuracy (permuted MNIST).}
    \label{fig:mnist_bwd}
  \end{subfigure}
 \caption{The prediction is over 7500 tasks and each data-point in the graph is averaged over 100 tasks for all models except for MAML. For MAML, we run over only 75 tasks and is  shown without averaging.}
  \label{fig:mnist_both}
  \vspace{-2mm}
\end{figure}

% ---------- SPLIT IMAGENET: forward + backward in one figure ----------
\subsection{Split Image Net}
\emph{Split-image-net} we use a support size of 180 for both query-only attention and full-attention model. For MAML a support size of only size 10 is enough.
\label{subsec:split-image-net}
\begin{figure}[H]
  \centering
  \begin{subfigure}[t]{0.48\linewidth}
    \centering
    \includegraphics[width=\linewidth]{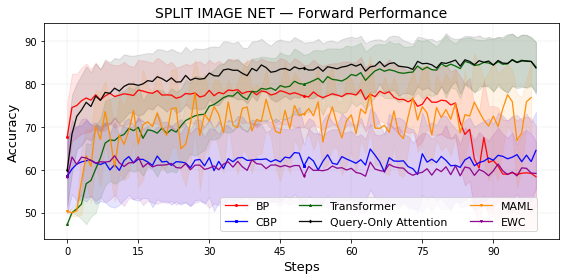}
    \caption{Forward accuracy (Split Image Net).}
    \label{fig:split_imagenet_fwd}
  \end{subfigure}\hfill
  \begin{subfigure}[t]{0.48\linewidth}
    \centering
    \includegraphics[width=\linewidth]{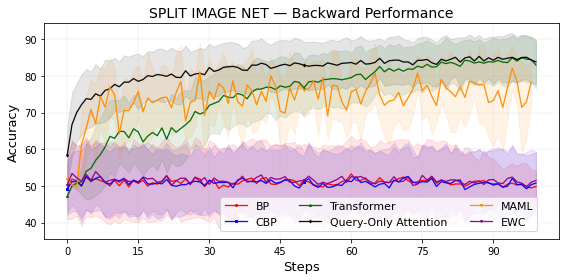}
    \caption{Backward accuracy (Split Image Net).}
    \label{fig:split_imagenet_bwd}
  \end{subfigure}
  \caption{The prediction is over 9000 tasks and each data-point in the graph is averaged over 100 tasks for all the models except MAML. MAML is run over 500 tasks, averaged over 5 tasks.  }
  \label{fig:split_imagenet_both}
  \vspace{-2mm}
\end{figure}
%\FloatBarrier

\paragraph{Observations in Classification Tasks.}
Across both \ref{subsec:permuted-mnist} and \ref{subsec:split-image-net}, the \textbf{query-only attention} model consistently outperforms all baselines in forward and backward performance. 
The \textbf{full-attention} reaches similar final accuracy but with 50\% more parameters and slower convergence. 
Under unknown task boundaries, \textbf{CBP} performs poorly, especially on Split ImageNet, while the \textbf{vanilla network} shows clear loss of plasticity. 
The \textbf{MAML}-based model converges fastest, achieving intermediate performance between attention-based and standard networks.

% ---------- SCR: forward + backward in one figure ----------
\subsection{Slowly Changing Regression}
In \emph{SCR}, we use a single query-only attention model with support size $100$, \emph{matching} the full-attention (both $100$) to isolate algorithmic effects under equal memory/compute.
With equal support ($100$) for query-only attention and full attention, both sustain low MSE in forward testing.
\label{subsec:SCR}
\begin{figure}[H]
  \centering
  \begin{subfigure}[t]{0.48\linewidth}
    \centering
    \includegraphics[width=\linewidth]{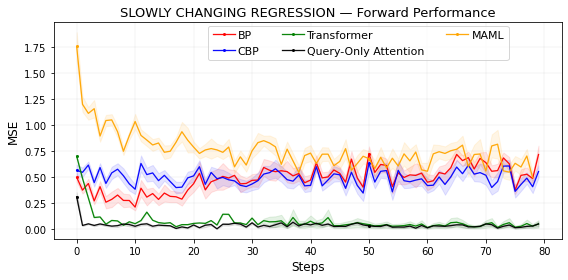}
    \caption{Forward MSE (SCR).}
    \label{fig:scr_fwd}
  \end{subfigure}\hfill
  \begin{subfigure}[t]{0.48\linewidth}
    \centering
    \includegraphics[width=\linewidth]{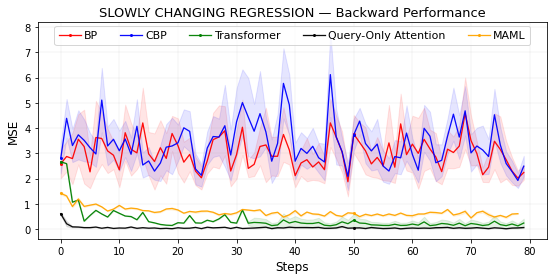}
    \caption{Backward MSE (SCR).}
    \label{fig:scr_bwd}
  \end{subfigure}
  \caption{The prediction is over 800 tasks and each data-point in the graph is averaged over 10 tasks for all models.}
  \label{fig:scr_both}
  \vspace{-2mm}
\end{figure}
%\FloatBarrier

\paragraph{Observations on Regression Task.}
As in classification, \textbf{BP} gradually loses plasticity, while \textbf{CBP} fails to learn effectively due to its purely online setup. 
The \textbf{query-only attention} model converges quickly with near-zero MSE, whereas \textbf{full attention} converges more slowly with slightly lower performance. 
Unlike in classification, the \textbf{MAML}-based model struggles to converge but still maintains plasticity. 
In backward testing, BP and CBP show severe forgetting, while query-only attention, full-attention, and MAML models retain high performance. 
Including $y_i$ in $Q_\theta(x_t, x_i, y_i)$ offered no gain on this regression task, so we used only $(x_t, x_i)$; for Permuted-MNIST, label inclusion improved results and was retained.

\paragraph{Result analysis.}
Across all benchmarks, \textbf{query-only attention} and \textbf{full-attention} models perform similarly, but the query-only attention model converges faster and scales better with $\mathcal{O}(n)$ complexity. 
The \textbf{MAML}-based approach shows strong backward performance and quick convergence with minimal support, though less consistent on SCR. 
\textbf{Query-only attention} model mitigates loss of plasticity and forgetting, highlighting its practicality under limited compute and memory.

\subsection{Effective Rank}
\label{subsec:effective-rank}

\begin{figure}[H]
  \centering
  % ---- Left figure ----
  \begin{subfigure}[t]{0.48\linewidth}
    \centering
    \includegraphics[width=\linewidth]{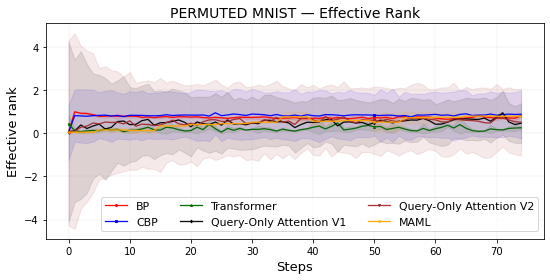}
    \label{fig:mnist_effective_rank}
  \end{subfigure}
  \hfill
  % ---- Right figure ----
  \begin{subfigure}[t]{0.48\linewidth}
    \centering
    \includegraphics[width=\linewidth]{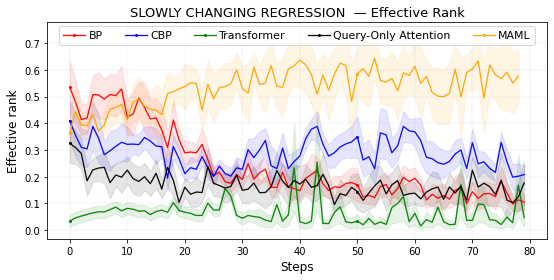}
    \label{fig:scr_effective_rank}
  \end{subfigure}
  \caption{ Effective rank co-varies with forward performance; dips align with reduced plasticity. Mean $\pm$ std over 3 seeds.}
  \label{fig:effective_rank_both}
  \vspace{-2mm}
\end{figure}

The effective rank serves as a proxy for plasticity.
We measure it at the start of each task using the Hessian of the final layer only, since full-Hessian computation is infeasible.
We measure it only on Permuted MNIST and SCR and not on image-net due to computational constraints. Also, the effective rank has been normalized since effective rank will depend on size of neural net. 
In both Permuted-MNIST and SCR, vanilla backpropagation shows a steady drop in effective rank, aligning with loss of plasticity. All other models show near minimal drop in effective rank thus indicating preserved plasticity.

\section{Conclusion}
\label{sec:conclusion}
We introduced a query-only attention mechanism for continual learning, showing that it mitigates both loss of plasticity when task boundary is unknown beating state of the art models and no task repetition. If the task boundary is known, then query-only attention can also mitigate catastrophic forgetting. Query-Only Attention has lower computational cost compared to full attention. Our analysis connects query-only models to MAML and full attention through the lens of global vs.\ local solutions, and further relates them to $k$-nearest neighbors. We further confirmed this relationship by running on three experiments.

Hessian rank experiments support the role of curvature in sustaining plasticity across different approaches which mitigate loss of plasticity.

A key limitation is the reliance on a support set, which complicates mitigation of catastrophic forgetting. Future work will extend to larger and more diverse benchmarks, more rigorous theoretical analysis and minimize the reliance on explicit support for the task at hand. These findings highlight that meta-learning principles, even in simplified architectures, can provide a path toward scalable continual learning.

\clearpage

\bibliographystyle{plainnat}
\bibliography{main}

%%%%%%%%%%%%%%%%%%%%%%%%%%%%%%%%%%%%%%%%%%%%%%%%%%%%%%%%%%%%

\appendix

\section{Technical Appendices and Supplementary Material}

\subsection{Broader impacts}
\label{sec:broader_impacts}
This work is primarily foundational research in continual learning. 
The proposed query-only attention mechanism and accompanying analysis aim to improve the 
understanding of plasticity and forgetting in neural networks. 
Potential positive impacts include enabling more efficient and adaptive AI systems, 
which could reduce retraining costs, improve energy efficiency, and support applications 
such as robotics, healthcare monitoring, and lifelong personal assistants.  

At the same time, continual learning technologies can be misused in domains such as 
surveillance or profiling, where adaptive models might amplify privacy concerns or biases.  
While the present work is not directly deployable, these risks highlight the need for 
responsible use and safeguards in future applications.  

Overall, this research contributes theoretical and empirical insights into the foundations 
of continual learning, with the aim of advancing the field in a transparent and beneficial direction.

\subsection{Limitations}
\label{sec:limitations}
Our work has a few important limitations.  
First, the query-only attention model relies on a support set, which can be restrictive for mitigating catastrophic forgetting in practical continual learning scenarios.  
Second, our evaluation is limited to two benchmarks (Permuted MNIST and Slowly Changing Regression); broader validation on more complex datasets is needed to confirm generality.  
Third, while we provide theoretical analysis and intuition linking query-only attention to MAML and $k$NN, we do not include formal proofs.  
We acknowledge these as areas for future work, particularly in extending the experimental scope and strengthening the theoretical foundation.

\subsection{Permuted MNIST Setup}
\label{sec:permuted_mnist_setup}
The MNIST dataset \citep{lecun1998gradient} consists of 60,000 training and 10,000 test images of hand-written digits ($0$–$9$), each represented as a $28 \times 28$ grayscale image.  
To adapt this dataset for continual learning, we make the following modifications:

\begin{itemize}
    \item \textbf{Train/test split.} For each task, we use the entire 60,000 original training images. These are further divided into 58,000 images for training and 2,000 images for evaluation within the task. The global MNIST test set is not used directly; instead, we re-sample 2,000 held-out examples per task to serve as test data. This ensures consistency across tasks and keeps evaluation lightweight.
    
    \item \textbf{Downsampling.} To reduce computational cost, all images are downsampled from $28 \times 28$ to $7 \times 7$, giving 49 input features per image.
    
    \item \textbf{Task generation.} Each task corresponds to a new random permutation of the 49 input pixels. The same permutation is applied consistently to all 60,000 images within a task. Labels remain unchanged.
    
    \item \textbf{Continual stream.} The learner observes tasks sequentially. After completing all 58,000 training pairs of a given permutation, the learner encounters remaining 2000 data-points for testing, following which it encounters the next task (with a new permutation). In total, 7,000 such tasks are generated in the continual stream.
\end{itemize}

This setup forces the continual learner to adapt to a new input representation (permutation) at the start of each task, while retaining performance on past permutations. It is a widely used benchmark for evaluating both \textit{plasticity} (ability to adapt to new tasks) and \textit{stability} (ability to avoid catastrophic forgetting).

\begin{table}[H]
\centering
\footnotesize
\caption{Permuted MNIST configuration: Query-Only Attention (V1).}
\label{tab:pmnist_query_only_attention_v1}
\begin{tabular}{@{}ll@{}}
\toprule
Setting & Value \\ \midrule
Support size ($|B|$) & 1000 \\
Replay buffer size & \texttt{1000} \\
Distance metric & Learned $Q_{\theta}$ (query-only) \\
Optimizer & Adam \\
Learning rate & \texttt{1e-4} \\
Weight decay & \texttt{0.0} \\
Batching & (batch size $=400$) \\
Seeds & 20, 30, 40 (report mean $\pm$ std) \\
Tasks & 7500 \\
Steps per task & \texttt{150} \\
(Input size, Hidden Size, Output Size, Hidden Layers) & ( 108, 100, 1, 9 ) \\ 
Init & Xavier uniform (weights), Zeros (biases) \\
Hardware & \texttt{1$\times$ RTX 3090 24GB, CUDA 11.8} \\
Wall-clock & \texttt{6.5 hours/run} \\
\bottomrule
\end{tabular}
\end{table}

\begin{table}[H]
\centering
\footnotesize
\caption{Permuted MNIST configuration: Query-Only Attention (V2).}
\label{tab:pmnist_query_only_attention_v2}
\begin{tabular}{@{}ll@{}}
\toprule
Setting & Value \\ \midrule
Support size ($|B|$) & 200 \\
Replay buffer size & \texttt{200} \\
Distance metric & Learned $Q_{\theta}$ (query-only) \\
Optimizer & Adam \\
Learning rate & \texttt{1e-4} \\
Weight decay & \texttt{0.0} \\
Batching & batch size $=400$ \\
Seeds & 20, 30, 40 (report mean $\pm$ std) \\
Tasks & 7500 \\
Steps per task & \texttt{150} \\
(Input size, Hidden Size, Output Size, Hidden Layers) & ( 108, 100, 1, 9 ) \\ 
Init & Xavier uniform (weights), Zeros (biases) \\
Hardware & \texttt{1$\times$ RTX 3090 24GB, CUDA 11.8} \\
Wall-clock & \texttt{4 hours/run} \\
\bottomrule
\end{tabular}
\end{table}

\begin{table}[H]
\centering
\footnotesize
\caption{Permuted MNIST configuration: MAML.}
\label{tab:pmnist_maml}
\begin{tabular}{@{}ll@{}}
\toprule
Setting & Value \\ \midrule
(Support size, query size, tasks per iteration) ($|B|$) & (10, 10, 5) \\
Replay buffer size & \texttt{50000} \\
Optimizer & Adam \\
Outer Learning rate & \texttt{1e-4} \\
Inner Learning rate & \texttt{1e-2} \\
Weight decay & \texttt{0.0} \\
Batching & batch size $=400$ \\
Seeds & 20, 30, 40 (report mean $\pm$ std) \\
Tasks & 75 \\
Steps per task & \texttt{150} \\
(Input size, Hidden Size, Output Size, Hidden Layers) : & ( 49, 100, 10, 3 ) \\ 
Init & Xavier uniform (weights), Zeros (biases) \\
Hardware & \texttt{1$\times$ RTX 3090 24GB, CUDA 11.8} \\
Wall-clock & \texttt{4.2 hours } \\
\bottomrule
\end{tabular}
\end{table}

\clearpage
The attention network for continual learning is directly adapted from \citep{wang2025replay}, which contains the comprehensive details.

\begin{table}[H]
\centering
\footnotesize
\caption{Permuted MNIST configuration: Attention Network baseline.}
\label{tab:pmnist_attention}
\begin{tabular}{@{}ll@{}}
\toprule
Setting & Value \\ \midrule
Support size ($|B|$) & 100 \\
Attention & Full self-attention ($\mathcal{O}(n^2)$) \\
Replay buffer size & \texttt{100} \\
Optimizer & Adam \\
Learning rate & \texttt{5e-4} \\
Weight decay & \texttt{0.0} \\
Batching & batch size $=400$ \\
Seeds & 20, 30, 40 (report mean $\pm$ std) \\
Tasks & 7500 \\
Steps per task & \texttt{150} \\
Architecture & \texttt{10 layers, 1 head, d\_model=59} \\
(Attention Layers, Attention Heads, Dimension) : & ( 10, 1, 59 ) \\ 
Init & Xavier uniform (weights), Zeros (biases) \\
Hardware & \texttt{1$\times$ RTX 3090 24GB, CUDA 11.8} \\
Wall-clock & \texttt{10 hours/run} \\
\bottomrule
\end{tabular}
\end{table}

The vanilla backpropagation and continual backpropagation algorithm is from paper \citep{dohare2024loss}, which contains more details.

\begin{table}[H]
\centering
\footnotesize
\caption{Permuted MNIST configuration: Vanilla Backpropagation.}
\label{tab:pmnist_bp}
\begin{tabular}{@{}ll@{}}
\toprule
Setting & Value \\ \midrule
Support size & N/A \\
Replay buffer size & 0 (no replay) \\
Optimizer & Adam \\
Learning rate & \texttt{1e-4} \\
Weight decay & \texttt{0.0} \\
Batching & Online (batch size $=1$) \\
Seeds & 20, 30, 40 (report mean $\pm$ std) \\
Tasks & 7500 \\
Steps per task & \texttt{150} \\
(Input size, Hidden Size, Output Size, Hidden Layers) : & ( 49, 200, 10, 3 ) \\ 
Init &  Xavier uniform (weights), Zeros (biases) \\
Hardware & \texttt{1$\times$ RTX 3090 24GB, CUDA 11.8} \\
Wall-clock & \texttt{5.0 hours/run} \\
\bottomrule
\end{tabular}
\end{table}

\begin{table}[H]
\centering
\footnotesize
\caption{Permuted MNIST configuration: Continuous Backpropagation.}
\label{tab:pmnist_bp}
\begin{tabular}{@{}ll@{}}
\toprule
Setting & Value \\ \midrule
Support size & N/A \\
Replay buffer size & 0 (no replay) \\
Optimizer & Adam  \\
Learning rate & \texttt{1e-4} \\
Weight decay & \texttt{0} \\
Batching & Online (batch size $=1$) \\
Seeds & 20, 30, 40 (report mean $\pm$ std) \\
Tasks & 7500 \\
Steps per task & \texttt{150} \\
(Input size, Hidden Size, Output Size, Hidden Layers) : & ( 49, 200, 10, 3 ) \\ 
Init &  Xavier uniform (weights), Zeros (biases)\\
Hardware & \texttt{1$\times$ RTX 3090 24GB, CUDA 11.8} \\
Wall-clock & \texttt{7.0 hours/run} \\
\bottomrule
\end{tabular}
\end{table}

\clearpage
\subsection{Split Image Net}
\label{sec:split)image_net_setup}
We adopt the Split Image Net benchmark introduced by Dohare et al.~\citep{dohare2024loss}. However, due to computational constraints we utilize 
tiny version of the image net instead of full image net. We further downsample the images to 32 * 32 for faster computation.
The remaining setup is same as in the original paper, ensuring comparability of results.  
For full details, we refer readers to the experimental protocol in \citep{dohare2024loss}.
Tiny image net consits of 200 labels and 500 images per label in training setup.
The test setup consists of 50 labels per class. 
To incorporate tiny image net for continual learning, each task consits of randomly sampling images from 2 classes. Thus a task is a binary classification task with 1000 datapoints for training. At the end of training in that class we measure the accuray on 100 datapoints corresponding to 2 lables used in training.
Since the training is purely online fashion a task is trained for a single epoch and then validation is carried out, followed by images for next task.
All the below architectures presented use the same CNN architure as the starting point for input image transformation:

\begin{table}[H]
\centering
\footnotesize
\caption{Split Image Net: CNN architecture (same for all models)}
\label{tab:split_image_net_cnn}
\begin{tabular}{@{}ll@{}}
\toprule
Setting & Value \\ \midrule
(Input channels, Output Channels,Kernel Size, Padding, MaxPool, Activation) & ( 3, 32, 5, 1, 2, Relu ) \\ 
(Input channels, Output Channels,Kernel Size, Padding, MaxPool, Activation) & ( 32, 64, 3, 1, 2, Relu ) \\ 
(Input channels, Output Channels,Kernel Size, Padding, MaxPool, Activation) & ( 64, 128, 3, 1, 2, Relu ) \\ 
Init & Xavier uniform (weights), Zeros (biases) \\
\bottomrule
\end{tabular}
\end{table}

\begin{table}[H]
\centering
\footnotesize
\caption{Split Image Net: Query-Only Attention.}
\label{tab:split_image_net_query_only_attention_v1}
\begin{tabular}{@{}ll@{}}
\toprule
Setting & Value \\ \midrule
Support size ($|B|$) & 180 \\
Replay buffer size & \texttt{200} \\
Distance metric & Learned $Q_{\theta}$ (query-only) \\
Optimizer & Adam \\
Learning rate & \texttt{1e-4} \\
Weight decay & \texttt{0.0} \\
Batching & (batch size $=10$) \\
Seeds & 20, 30, 40 (report mean $\pm$ std) \\
Tasks & 9500 \\
Steps per task & \texttt{150} \\
(Input size, Hidden Size, Output Size, Hidden Layers) & ( 2304, 128, 1, 1 ) \\ 
Init & Xavier uniform (weights), Zeros (biases) \\
Hardware & \texttt{1$\times$ RTX 3090 24GB, CUDA 11.8} \\
Wall-clock & \texttt{3 hours/run} \\
\bottomrule
\end{tabular}
\end{table}

\begin{table}[H]
\centering
\footnotesize
\caption{Split Image Net: MAML.}
\label{tab:split_image_net_maml}
\begin{tabular}{@{}ll@{}}
\toprule
Setting & Value \\ \midrule
(Support size, query size, tasks per iteration) ($|B|$) & (10, 10, 5) \\
Replay buffer size & \texttt{20000} \\
Optimizer & Adam \\
Outer Learning rate & \texttt{1e-4} \\
Inner Learning rate & \texttt{1e-2} \\
Weight decay & \texttt{0.0} \\
Batching & batch size $=10$ \\
Seeds & 20, 30, 40 (report mean $\pm$ std) \\
Tasks & 500 \\
Steps per task & \texttt{150} \\
(Input size, Hidden Size, Output Size, Hidden Layers) : & ( 1152, 128, 2, 1 ) \\ 
Init & Xavier uniform (weights), Zeros (biases) \\
Hardware & \texttt{1$\times$ RTX 3090 24GB, CUDA 11.8} \\
Wall-clock & \texttt{5 hours (ran only first 100 tasks)} \\
\bottomrule
\end{tabular}
\end{table}

\begin{table}[H]
\centering
\footnotesize
\caption{Split Image Net: Attention Network baseline.}
\label{tab:split_image_net_attention}
\begin{tabular}{@{}ll@{}}
\toprule
Setting & Value \\ \midrule
Support size ($|B|$) & 180 \\
Attention & Full self-attention ($\mathcal{O}(n^2)$) \\
Replay buffer size & \texttt{200} \\
Optimizer & Adam \\
Learning rate & \texttt{1e-4} \\
Weight decay & \texttt{0.0} \\
Batching & batch size $=10$ \\
Seeds & 20, 30, 40 (report mean $\pm$ std) \\
Tasks & 9500 \\
Steps per task & \texttt{150} \\
Architecture & \texttt{3 layers, 1 head, d\_model=130} \\
Init & Xavier uniform (weights), Zeros (biases) \\
Hardware & \texttt{1$\times$ RTX 3090 24GB, CUDA 11.8} \\
Wall-clock & \texttt{4.5 hours/run} \\
\bottomrule
\end{tabular}
\end{table}

The vanilla backpropagation and continual backpropagation algorithm is from paper \citep{dohare2024loss}, which contains more details.

\begin{table}[H]
\centering
\footnotesize
\caption{Split Image Net: Vanilla Backpropagation.}
\label{tab:split_image_net_bp}
\begin{tabular}{@{}ll@{}}
\toprule
Setting & Value \\ \midrule
Support size & N/A \\
Replay buffer size & 0 (no replay) \\
Optimizer & Adam \\
Learning rate & \texttt{1e-4} \\
Weight decay & \texttt{0.0} \\
Batching & Online (batch size $=10$) \\
Seeds & 20, 30, 40 (report mean $\pm$ std) \\
Tasks & 9500 \\
Steps per task & \texttt{150} \\
(Input size, Hidden Size, Output Size, Hidden Layers) : & ( 1152, 128, 2, 1 ) \\ 
Init &  Xavier uniform (weights), Zeros (biases) \\
Hardware & \texttt{1$\times$ RTX 3090 24GB, CUDA 11.8} \\
Wall-clock & \texttt{4.0 hours/run} \\
\bottomrule
\end{tabular}
\end{table}

\begin{table}[H]
\centering
\footnotesize
\caption{Split Image Net: Continuous Backpropagation.}
\label{tab:split_image_net_cbp}
\begin{tabular}{@{}ll@{}}
\toprule
Setting & Value \\ \midrule
Support size & N/A \\
Replay buffer size & 0 (no replay) \\
Optimizer & Adam  \\
Learning rate & \texttt{1e-4} \\
Weight decay & \texttt{0} \\
Batching & Online (batch size $=1$) \\
Seeds & 20, 30, 40 (report mean $\pm$ std) \\
Tasks & 7000 \\
Steps per task & \texttt{150} \\
(Input size, Hidden Size, Output Size, Hidden Layers) : & ( 1152, 128, 2, 1 ) \\ 
Init &  Xavier uniform (weights), Zeros (biases)\\
Hardware & \texttt{1$\times$ RTX 3090 24GB, CUDA 11.8} \\
Wall-clock & \texttt{5.5 hours/run} \\
\bottomrule
\end{tabular}
\end{table}

\clearpage

\subsection{Slowly changing regression (SCR)}
\label{sec:scr_setup}
We adopt the Slowly changing regression (SCR) benchmark introduced by Dohare et al.~\citep{dohare2024loss}, 
which was designed to study loss of plasticity in continual learning.  
In this task, regression targets evolve gradually over time according to smoothly drifting functions, 
creating a non-stationary data stream.  
We use the same setup and data-generation procedure as in the original paper, ensuring comparability of results.  
For full details, we refer readers to the experimental protocol in \citep{dohare2024loss}.

\begin{table}[H]
\centering
\footnotesize
\caption{Slowly changing regression task: Query-Only Attention.}
\label{tab:scr_query_only_attention}
\begin{tabular}{@{}ll@{}}
\toprule
Setting & Value \\ \midrule
Support size ($|B|$) & 100 \\
Replay buffer size & \texttt{100} \\
Distance metric & Learned $Q_{\theta}$ (query-only) \\
Optimizer & Adam \\
Learning rate & \texttt{1e-4} \\
Weight decay & \texttt{0.0} \\
Batching &  batch size $=1$ \\
Seeds & 20, 30, 40 (report mean $\pm$ std) \\
Tasks & 800 \\
Steps per task & \texttt{10000} \\
(Input size, Hidden Size, Output Size, Hidden Layers) : & ( 40, 20, 1, 1 ) \\ 
Init & Xavier uniform (weights), Zeros (biases)\\
Hardware & \texttt{1$\times$ RTX 3090 24GB, CUDA 11.8} \\
Wall-clock & \texttt{7 hours/run} \\
\bottomrule
\end{tabular}
\end{table}

\begin{table}[H]
\centering
\footnotesize
\caption{Slowly changing regression task: MAML.}
\label{tab:scr_maml}
\begin{tabular}{@{}ll@{}}
\toprule
Setting & Value \\ \midrule
(Support size, query size, tasks per iteration) ($|B|$) & (10, 10, 5) \\
Replay buffer size & \texttt{20000} \\
Optimizer & Adam \\
Inner Learning rate & \texttt{1e-2} \\
Outer Learning rate & \texttt{1e-4} \\
Weight decay & \texttt{0.0} \\
Batching & batch size $=1$ \\
Seeds & 20, 30, 40 (report mean $\pm$ std) \\
Tasks & 800 \\
Steps per task & \texttt{10000} \\
(Input size, Hidden Size, Output Size, Hidden Layers) : & ( 20, 40, 1, 1 ) \\ 
Init & Xavier uniform (weights), Zeros (biases)\\
Hardware & \texttt{1$\times$ RTX 3090 24GB, CUDA 11.8} \\
Wall-clock & \texttt{8.5 hour to run 100 tasks} \\
\bottomrule
\end{tabular}
\end{table}

\begin{table}[H]
\centering
\footnotesize
\caption{Slowly changing regression task: Attention Network.}
\label{tab:scr_attention}
\begin{tabular}{@{}ll@{}}
\toprule
Setting & Value \\ \midrule
Support size ($|B|$) & 100 \\
Replay buffer size & \texttt{100} \\
Distance metric & Learned $Q_{\theta}$ (query-only) \\
Optimizer & Adam \\
Learning rate & \texttt{1e-4} \\
Weight decay & \texttt{0.0} \\
Batching & batch size $=1$ \\
Seeds & 20, 30, 40 (report mean $\pm$ std) \\
Tasks & 800 \\
Steps per task & \texttt{10000} \\
(Attention Layers, Attention Heads, Dimension) : & ( 1, 1, 21 ) \\ 
Init & Xavier uniform (weights), Zeros (biases)\\
Hardware & \texttt{1$\times$ RTX 3090 24GB, CUDA 11.8} \\
Wall-clock & \texttt{14 hours/run} \\
\bottomrule
\end{tabular}
\end{table}

\begin{table}[H]
\centering
\footnotesize
\caption{Slowly changing regression task: Vanilla Backpropagation}
\label{tab:scr_bp}
\begin{tabular}{@{}ll@{}}
\toprule
Setting & Value \\ \midrule
Optimizer & Adam \\
Learning rate & \texttt{1e-4} \\
Weight decay & \texttt{0.0} \\
Batching & batch size $=1$ \\
Seeds & 20, 30, 40 (report mean $\pm$ std) \\
Tasks & 800 \\
Steps per task & \texttt{10000} \\
(Input size, Hidden Size, Output Size, Hidden Layers) : & ( 20, 40, 1, 1 ) \\ 
Init & Xavier uniform (weights), Zeros (biases)\\
Hardware & \texttt{1$\times$ RTX 3090 24GB, CUDA 11.8} \\
Wall-clock & \texttt{8 hours/run} \\
\bottomrule
\end{tabular}
\end{table}

\begin{table}[H]
\centering
\footnotesize
\caption{Slowly changing regression task: Continuous Backpropagation}
\label{tab:scr_cbp}
\begin{tabular}{@{}ll@{}}
\toprule
Setting & Value \\ \midrule
Optimizer & Adam \\
Learning rate & \texttt{1e-4} \\
Weight decay & \texttt{0.0} \\
Batching & batch size $=1$ \\
Seeds & 20, 30, 40 (report mean $\pm$ std) \\
Tasks & 800 \\
Steps per task & \texttt{10000} \\
(Input size, Hidden Size, Output Size, Hidden Layers) : & ( 20, 40, 1, 1 ) \\ 
Init & Xavier uniform (weights), Zeros (biases)\\
Hardware & \texttt{1$\times$ RTX 3090 24GB, CUDA 11.8} \\
Wall-clock & \texttt{11 hours/run} \\
\bottomrule
\end{tabular}
\end{table}

\end{document}